\documentclass{INTERSPEECH2023}

\interspeechcameraready

\usepackage{graphicx, caption, subcaption}
\usepackage{multirow}
\usepackage{comment}
\usepackage{adjustbox}

\title{MMViT: Multiscale Multiview Vision Transformers}
\name{Yuchen Liu$^{1,2}$, Natasha Ong$^{1}$, Kaiyan Peng$^{1}$, Bo Xiong$^{1}$, Qifan Wang$^{1}$, Rui Hou$^{1}$, Madian Khabsa$^{1}$, Kaiyue Yang$^{1}$, David Liu$^{1}$,  Donald S. Williamson$^{3}$, Hanchao Yu$^{1}$}

\address{$^1$ Meta Platforms Inc., $^2$ Indiana University Bloomington, $^3$ The Ohio State University}
\email{liu477@iu.edu, natashasaki@meta.com, kaiyanp@meta.com, bxiong@meta.com, wqfcr@meta.com, rayhou@meta.com, mkhabsa@meta.com, kaiyue@meta.com, davidportal@meta.com, williamson.413@osu.edu, hanchaoyu@meta.com}

\begin{document}
\maketitle
\begin{abstract}

We present Multiscale Multiview Vision Transformers (MMViT), which introduces multiscale feature maps and multiview encodings to transformer models. Our model encodes different views of the input signal and builds several channel-resolution feature stages to process the multiple views of the input at different resolutions in parallel. At each scale stage, we use a cross-attention block to fuse information across different views. This enables the MMViT model to acquire complex high-dimensional representations of the input at different  resolutions. The proposed model can serve as a backbone model in multiple domains. We demonstrate the effectiveness of MMViT on audio and image classification tasks, achieving state-of-the-art results.

\end{abstract}

\section{Introduction}
In recent years, convolutional neural networks have been the dominant models for representation learning 
across multiple domains~\cite{he2016deep, simonyan2014very, cramer2019look, arandjelovic2017look, gong2021psla}. However, recent research has demonstrated a significant shift towards transformer-based architectures for representation learning~\cite{devlin2018bert, dosovitskiy2020image, xu2022masked, chen2022beats,hsu2021hubert}. Transformers, introduced in~\cite{vaswani2017attention}, have shown superior performance and scalability with large-scale data. 
Transformers utilize self-attention mechanisms to effectively capture long-range dependencies and extract contextual information from the input sequence. Therefore, 
transformers are more suitable for tasks that require modeling of complex relationships and dependencies among the input features.

The recent shift towards transformer-based architectures for representation learning has been driven by several breakthroughs in the field, such as the success of transformer-based models like Vision-Transformer (ViT)~\cite{dosovitskiy2020image} in computer vision. This concept was quickly applied to audio as well~\cite{gong2021ast}. Researchers have built on Vision Transformers by adding several promising upgrades. One class of upgrades is the Multiscale Vision Transformer (MViT)~\cite{fan2021multiscale}, which has been shown to build multiscale feature hierarchies with transformer models. The multiscale pyramid of features employed by MViT enable the model to progressively change the spatio-temporal resolution of the input features, expanding feature complexity while reducing visual resolution. This approach allows MViT models to achieve better performance on both video and image recognition tasks while also reducing the inference cost in terms of FLOPs. Another class of upgrades that researchers have explored is the multiview vision transformer. An example of such a model is the Multiview Transformers for Video Recognition (MTV)~\cite{yan2022multiview} model developed by Google. Unlike the pyramid-based approach, the MTV model employs separate encoders to represent different views of the input video, and uses lateral connections to fuse information across views. The different views are generated by tokenizing the video using tubelets of varying sizes. These multiview models have shown promising results in video recognition tasks, demonstrating the potential for transformer-based architectures to handle complex spatio-temporal relationships.

We present a novel transformer-based model called Multiscale Multiview Vision Transformers (MMViT) that combines the strengths of the MViT and MTV models. Unlike the original MViT model, which only applies self-attention on the input, MMViT improves upon it by introducing a cross-attention layer at each scale stage to merge information from multiple views. This enables the MMViT model to acquire multi-resolution temporal context at each scale stage by processing multiple views of the input at different resolutions in parallel. Additionally, the MMViT model uses a hierarchical scaling system to increase channel size and reduce spatial resolution, generating high-dimensional complex features as the network deepens. By employing multiscale and multiview features, the MMViT model enhances its representation learning capability and overall performance. 

Our model is a versatile architecture that can be applied to 
multiple domains. In this paper, we focus on audio and image classification. 
For audio data, the input can be transformed into a 2D image representation (a spectrogram). On the other hand, images are already naturally represented as 2D images, making them compatible with the MMViT model. We begin by training and testing the MMViT model on image classification tasks using the widely-used ImageNet1K dataset~\cite{russakovsky2015imagenet}. After achieving competitive performance, we transfer the knowledge learned from the image model to the MMViT model for audio classification tasks using the Audioset dataset~\cite{gemmeke2017audio}. To prevent overfitting in the audio MMViT model, we propose an audio-specific CutMix method. Our results demonstrate that the MMViT model achieves state-of-the-art performance on both audio and image classification tasks, outperforming several baselines. Moreover, the multiscale and multiview features can effectively represent complex input signals in a more interpretable and efficient manner. This makes the proposed MMViT model applicable to other modalities, such as video and text. We believe that the MMViT model presents a new and effective approach to representation learning, which can be widely used in various downstream applications, including video recognition  and object detection.
\begin{figure*}[t]
  \centering
  \includegraphics[scale=0.16]{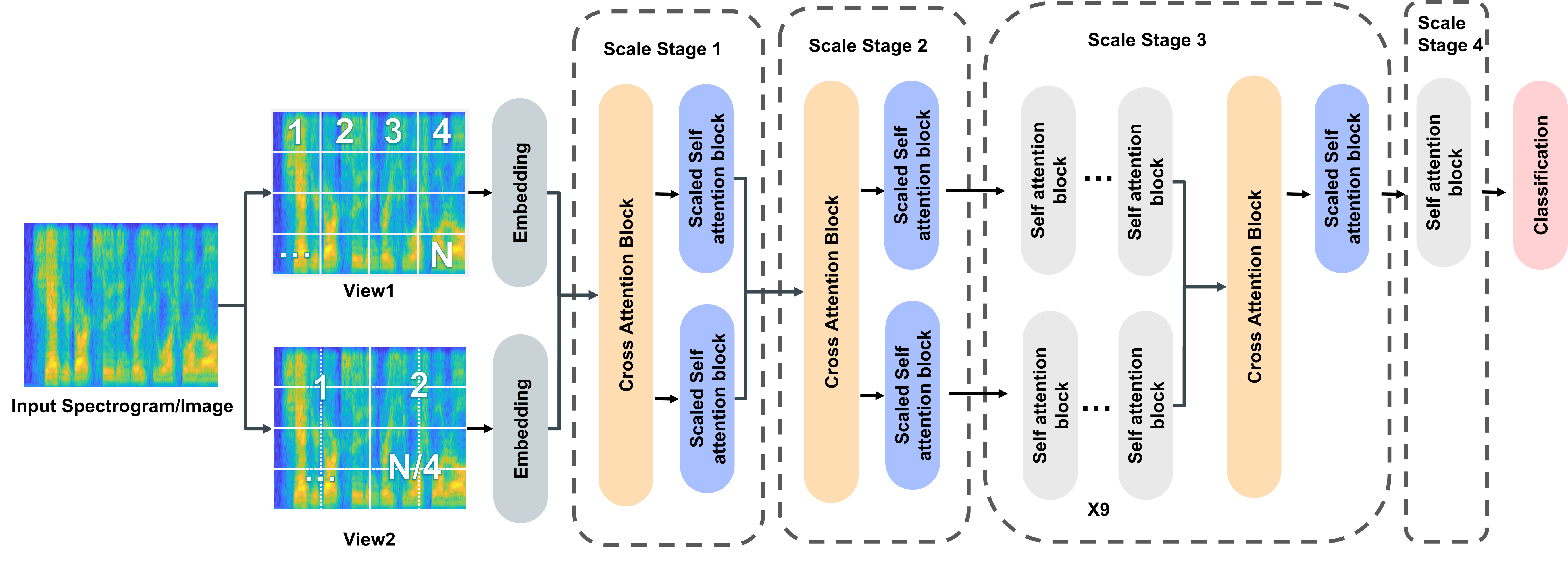}
  \caption{An overview of the 16-block MMViT model. 
  The input spectrogram is first patchified into two views with different resolutions. Both views are then encoded with a learnable spatio-temporal positional encoding layer. The embedded features pass through four scale stages, each with multiple transformer blocks. The second-to-last block of each stage except the last stage is used to fuses the information from both views using cross-attention, while the last block uses scaled self-attention to reduce spatio-temporal resolution and increase feature channels. The other blocks use regular self-attention.} 
  \label{fig:overview}
  \vspace{-5mm}
\end{figure*}
\section{Related Work}
Representation learning using transformer-based models becomes increasingly popular in various domains, including audio and image data. Initially, transformer-based models were mainly used in natural language processing~\cite{devlin2018bert, he2020deberta, yang2019xlnet, radford2018improving}. However, with the advent of Vision Transformer (ViT)~\cite{dosovitskiy2020image}, transformer-based models have been directly applied to image classification tasks by treating image patches as word embeddings. To further enhance the ViT's effectiveness with smaller datasets, DeiT~\cite{touvron2021training} proposed various training strategies. Similarly, Swin Transformer~\cite{liu2021swin} introduced a hierarchical Transformer architecture that uses shifted windows to further improve the model. The Multiscale Vision Transformers~\cite{fan2021multiscale} utilized multi-head pooling attention on video and image recognition tasks and achieved similar results. In contrast, Multiview Transformers~\cite{yan2022multiview} proposed a separate encoder to represent different views of the input and fuse the information at the same level horizontally through lateral connections, instead of scaling the input throughout the network.

The ViT approach has also been applied to the audio domain. Gong \textit{et al.} were among the first to use ViT for audio data by treating audio spectrograms as images and createt the Audio Spectrogram Transformer (AST) model~\cite{gong2021ast}. The study found that cross-modality transfer learning works well on audio tasks by using pre-trained models from computer vision and overlapping patches from audio spectrograms for fine-tuning. Khaled \textit{et al.} further optimized and regularized the AST model by using Patchout to reduce the computation and memory complexity of training transformers for audio~\cite{koutini2021efficient}. Another approach, called HTS-AT~\cite{chen2022hts}, used a hierarchical structure to decrease the size of the model and shorten the training time required by audio spectrogram transformers.

\section{Approach} \label{method}
In this section, we present an overview of the model architecture and experimental methodology used in the proposed MMViT model (see Figure \ref{fig:overview}). 

\subsection{Input and Embedding}
For both image and audio classification tasks, we represent the input as a two-dimensional image. For audio input, we utilize the log Mel filterbank features as the input image for MMViT. The input image dimension can be expressed in the form of $C \times H \times W$, where $C$ refers to the channel size. Specifically,  $C$ = 3 for image input and $C$ = 1 for audio input.

To generate distinct views of the input image at varying resolutions, we initially patchify the input images into two views using 2D convolutional operations. The final number of patches for each view is determined by the chosen stride during the convolution process. We double the stride for both height and width for the second view, resulting in the patch number for the second view being a quarter of the first view. For this study, the first view of the input data is convolved using a kernel size of [9,9] and a stride of [2,2]. The second view, on the other hand, uses a kernel size and stride of [13,13] and [4,4], respectively. It's worth noting that, unlike traditional ViT, our patchification process introduces overlap by utilizing a kernel size that is larger than stride. This approach ensures that the resulting patches encompass a greater amount of contextual information from surrounding pixels than traditional ViT. This procedure enables the MMViT model to capture and integrate information in parallel within subsequent transformer blocks, which is crucial for achieving high performance in image and audio classification tasks.

Subsequently, we apply a learnable spatio-temporal cls positional encoding\footnote{\url{https://pytorchvideo.readthedocs.io/en/latest/api/layers/layers.html}} to both generated views, with one positional encoding computed separately for spatial patches and temporal sequences. This positional encoding process helps the model encode positional information and overcome permutation invariance. Similar to other transformer-based embedding models~\cite{devlin2018bert}, we append a $\left[CLS\right]$ token solely to the beginning of the first view. This $\left[CLS\right]$ token serves as the input for the classification head to predict the classification outcome. 
As a result, the output size for the first view following the embedding layer is $E \times\left(\frac{H}{2} \times \frac{W}{2}+1\right)$, where $E$ denotes the embedding size and the addition of $+1$ accounts for the $\left[CLS\right]$ token. In contrast, for view 2, the output size after patchification and positional embedding process is $E \times \frac{H}{4} \times \frac{W}{4}$.
\begin{figure}[t]
    \centering
    \begin{subfigure}[t]{0.3\textwidth}
        \includegraphics[width=\linewidth]{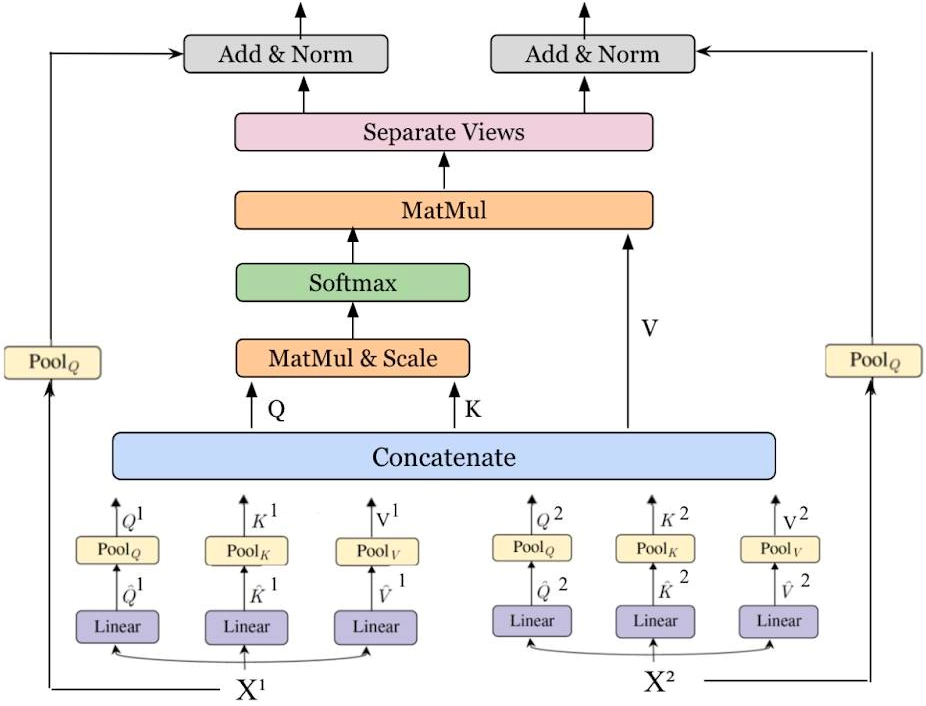}
        \caption{Cross attention block}
        \label{fig:cross}
    \end{subfigure}
    \begin{subfigure}[t]{0.3\textwidth}
         \includegraphics[width=\linewidth]{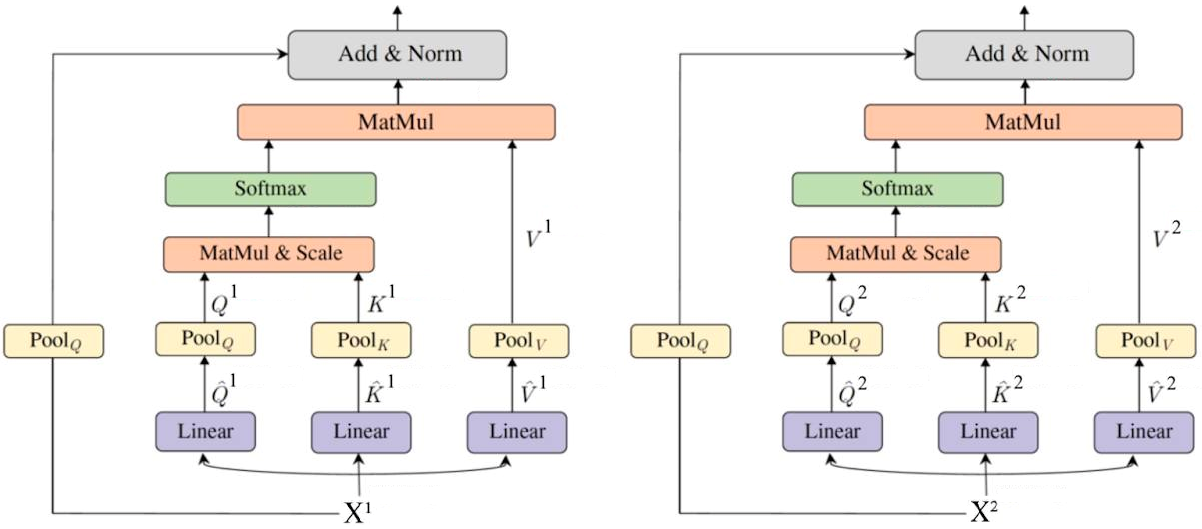}
         \caption{Self attention block}
         \label{fig:self}
    \end{subfigure}
\caption{MMViT blocks structure for a) cross attention block which fuses information between two views at the same stage; b) self-attention layer for two views with multi-head pooling attention equivalent to a MViT block~\cite{fan2021multiscale}.}
\label{fig:block}
\vspace{-5mm}
\end{figure}
\subsection{Scale Stage}
Similar to MViT, the scale stage in MMViT refers to a group of transformer blocks that function at the same scale and possess identical resolution across space-temporal dimensions and channels. Our proposed MMViT architecture comprises four scale stages and a total of 16 layers of transformer blocks with 2 views processed in each layer. At each scale stage, $n$ self-attention blocks are followed by a cross-attention block and a scaled self-attention block. The value of $n$ are [0,0,9,1] for each scale stage, respectively. The space-temporal dimensions and channel size are only modified at scaled self-attention blocks. To summarize, the proposed MMViT model is composed of cross-attention blocks located at layers 1, 3, and 14, scaled self-attention blocks located at layers 2, 4, and 15, and self-attention blocks situated at layers 5-13 and 16. A detailed structural breakdown of each block is provided in Figure \ref{fig:block}.

\subsubsection{Self-attention Block}  Each self-attention layer in our architecture generates identical self-attention blocks same as the number of views showed in Figure \ref{fig:self}, which is equivalent to the MViT block described in~\cite{fan2021multiscale}. Each of these blocks corresponds to a particular view, and each view is processed separately through its own transformer tower using Multi-Head Pooling Attention (MHPA). To begin with, the input view $X_{i}$ generates its own  $\hat{Q}_{i}$, $\hat{K}_{i}$, $\hat{V}_{i}$ pair through separate linear layers. The pooling operator is then applied to each pair to produce ${Q}_{i}$, ${K}_{i}$, ${V}_{i}$. The pooling operation $P=(\cdot ; \theta)$ is implemented using a convolutional operator with a parameter set of $(kernel, stride, padding)$ and can be expressed as:

\begin{equation}
\begin{split}
& \operatorname{Pooling Attention}(\cdot)=  \\
& \operatorname{Softmax}\left(\mathcal{P}\left({Q_i} ; \theta_{Q_i}\right) \mathcal{P}\left(K_{i} ; \theta_{K_i}\right)^T \sqrt{d}\right)  \mathcal{P}\left({V_i} ; \theta_{V_i}\right)
\end{split}
\end{equation}

At the last block of each scale stage except the last stage, the scaled self-attention block uses the $\mathcal{P}\left(Q_i ; \boldsymbol{\theta}_{Q_i}\right)$ operation with a stride of $[2,2]$ to downsample the spatio-temporal dimension of the feature $X_i$ by a factor of four. Additionally, the channel size is doubled by applying multiple linear layers at the end of the scaled self-attention block.

\subsubsection{Cross-attention Block} The cross-attention blocks are utilized to merge information from different views at each scale stage. To take advantage of the most learned information in the views prior to scaling down to lower resolutions, the cross-attention blocks are positioned before the scale self-attention block. The structure of the cross-attention blocks is presented in detail in Figure \ref{fig:cross}. Similar to the self-attention block, the input views are first passed through linear layers and pooling attention layer to generate ${Q}_{i}$, ${K}_{i}$, ${V}_{i}$ for each view. Subsequently, we concatenate all ${Q}_{i}$, ${K}_{i}$, ${V}_{i}$ across spatial and temporal dimensions to form $Q, K, V$ for cross-attention. The cross pooling attention is expressed as:

\begin{equation}
\begin{split}
& \operatorname{Cross Pooling Attention}(\cdot)=  \\
& \operatorname{Softmax}\left(\prod_{i=0}^N \mathcal{P}\left(Q_i ; \theta_{Q_i}\right) \prod_{i=0}^N \mathcal{P}\left(K_i ; \theta_{K_i}\right)^T \sqrt{d}\right)  \prod_{i=0}^N \mathcal{P}\left(V_i ; \theta_{V_i}\right)
\end{split}
\end{equation}

where $\prod_{i=0}^N$ denotes the concatenation operation and $N$ is the number of views. The cross-attention mechanism is utilized to extract global context information and merge the information between each view. Following the attention mechanism, the output feature is separated for each view for future blocks.

\section{Experiment Setup} 
\subsection{Datasets}
In our audio classification experiment, we utilize both the balanced and unbalanced versions of the AudioSet dataset~\cite{gemmeke2017audio}. AudioSet is a collection of over 2 million 10-second audio clips extracted from YouTube videos and labeled with one or more of 527 sound classes. The balanced version of AudioSet contains over 22k samples with labels that have similar weights, while the unbalanced/full version contains over 2 million samples, and its evaluation set consists of 20k samples.

For our image classification task, we employ the renowned ImageNet 1K dataset~\cite{russakovsky2015imagenet}. This dataset is a large-scale collection of labeled images that is extensively used to train and evaluate computer vision models. The ImageNet 1K dataset comprises 1.2 million training images and 50,000 validation images, categorized into 1,000 different object classes.

\subsection{Training Details}
\subsubsection{Training Setup}
In the audio experiment, the input audio waveform is converted into a sequence of 128-dimensional log Mel fbank features computed using a 25ms Hamming window with a 10ms shift. Similar to Gong \textit{et al.}~\cite{gong2021ast}, weighted sampling is used for the full Audioset dataset. The AdamW~\cite{loshchilov2017decoupled} optimizer is employed with weight decay of $1\mathrm{e}{-4}$, and an initial learning rate of $1\mathrm{e}{-5}$. The model is trained for 150 epochs on the audioset balanced set and 100 epochs on the audioset full set, with mean average precision (mAP) used as the evaluation metric.

In the image experiment, the AdamW optimizer with weight decay of $1\mathrm{e}{-2}$ is used, with an initial learning rate of $5\mathrm{e}{-5}$. The model is trained for 400 epochs on the ImageNet 1K dataset, and the evaluation is done using Top1 accuracy metrics.

\subsubsection{Data Augmentations} \label{sec:dataaug}

\begin{figure}[t]
  \centering
  \includegraphics[scale=0.6]{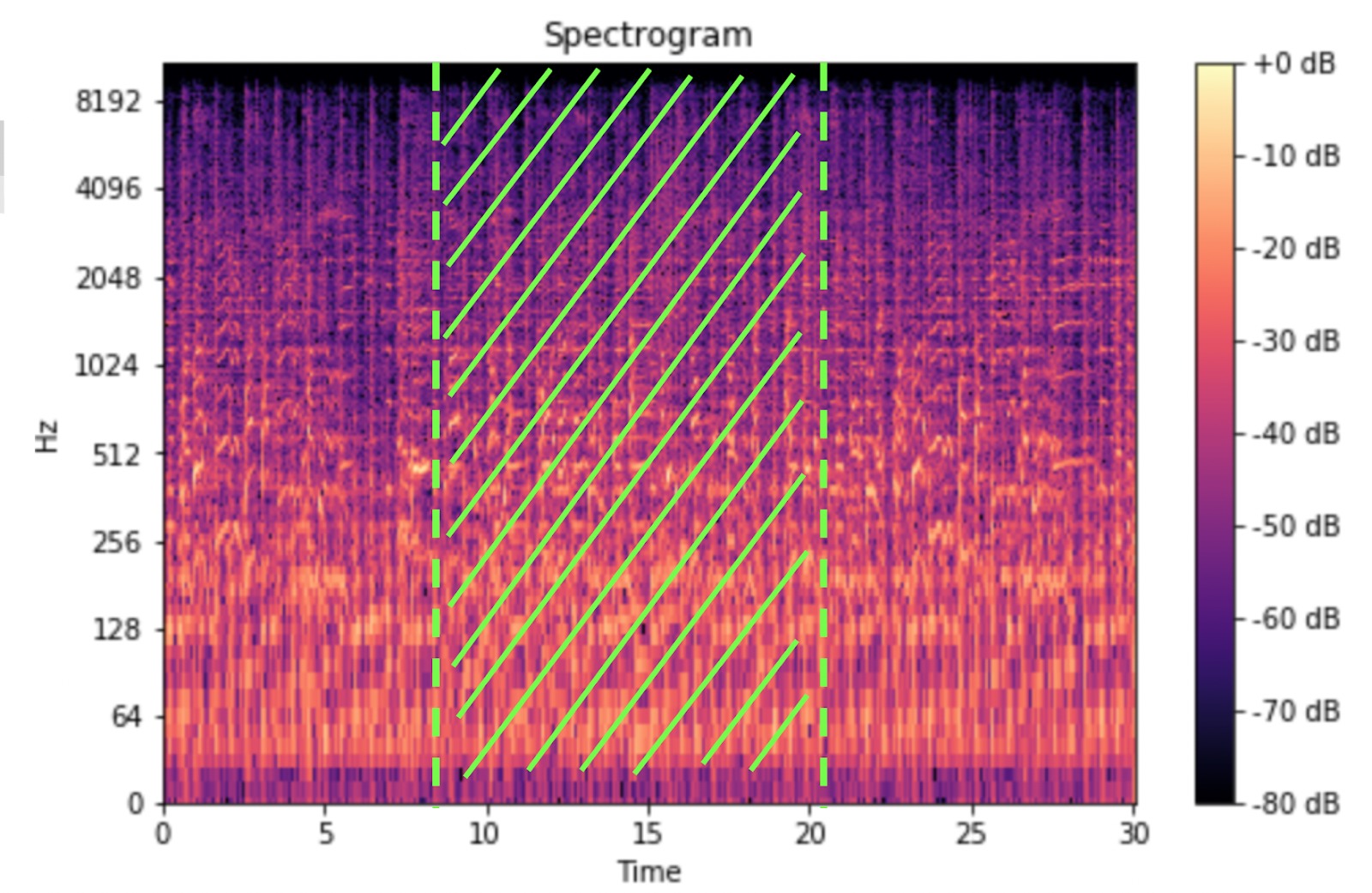}
  \caption{Audio-specific Cutmix which contains only a temporal axis cut.} 
  \label{fig:cutmix}
  \vspace{-5mm}
\end{figure}

In transformer-based audio representation learning, audio data augmentation is essential as the model is prone to overfitting. Our audio experiments include various augmentation techniques including Mixup~\cite{zhang2018mixup} and Cutmix~\cite{yun2019cutmix}, which are commonly used in audio data augmentation\cite{gong2021ast,koutini2021receptive}. Additionally, we introduce an audio-specific data augmentation method called audio Cutmix as shown in Figure \ref{fig:cutmix}. Audio cutmix is similar to image Cutmix~\cite{yun2019cutmix}, but has an temporal axis cut because the frequency axis contains quantitative information that is not invariant as in images. Throughout the entire experiment training process, including the process for training the baseline models, half of the audio clips are augmented using Mixup and the other half are augmented using our audio Cutmix technique.

We also utilize the Specaugment~\cite{park19e_interspeech}, which involves applying spectrogram masking with a maximum time mask length of 192 frames and a maximum frequency mask length of 48 bins. Another augmentation technique we use is random rolling, which randomly shifts each audio clip in time by a certain number of samples. This random shift causes the entire waveform to appear as if it was played earlier or later in time, which prevents the model from overfitting to the training data.

We adopted the identical data augmentation pipeline as MViT~\cite{fan2021multiscale} for our image classification experiment, which encompasses Mixup~\cite{zhang2018mixup}, Cutmix~\cite{yun2019cutmix}, Random erasing~\cite{zhong2020random}, and Randaugment~\cite{cubuk2020randaugment}.

\subsection{Pre-training}
 Gong \textit{et al.}~\cite{gong2021ast} demonstrated that leveraging pretrained models on the ImageNet dataset can significantly enhance the performance of audio transformer models on the AudioSet dataset. This is achieved by reducing the need for large amounts of in-domain audio data. We adopt the same strategy and use ImageNet1K pretrained models in our audio experiments.

For the patchification step, we average the weights in the 3 image channels to 1 channel for audio. For the learnable positional embedding weights and other MMViT model weights, we simply interpolate the weights from the image MMViT model to the audio MMViT model. We directly reuse the positional embedding for the $\left[CLS\right]$ token. Finally, we remove the image classification head and re-initialize it for the audio experiment.

\section{Results}
\subsection{Audio Classification Results}

\begin{table}[]
\centering
\caption{Audio classification results.}
\begin{tabular}{l|l|c}
\hline
Model                                      & Dataset & \multicolumn{1}{l}{mAP}        \\ \hline
AST~\cite{gong2021ast}     & balance & 31.8                           \\
MViTv2~\cite{li2022mvitv2} & balance & 32                             \\
MMViT                                      & balance & \textbf{32.2} \\ \hline
AST~\cite{gong2021ast}     & full    & 37.2                           \\
MViTv2~\cite{li2022mvitv2} & full    & 42.4                           \\
MMViT                                      & full    & \textbf{43}   \\
\hline
\end{tabular}
\label{tab:audio_res}

\end{table}


We show audio classification results on both balanced and full Audioset in Table \ref{tab:audio_res}.
 We use two baselines: 1) Audio Spectrogram Transformer~\cite{gong2021ast} \footnote{The results presented in the table for these baselines were obtained after retraining them based on the Audio-MAE codebase from Facebook Research at \url{https://github.com/facebookresearch/AudioMAE}. As a result, the final outcomes may differ slightly from those reported in the original paper.}; and 2) the improved Multiscale Vision Transformers (MViTv2)~\cite{li2022mvitv2}. Note that MViTv2~\cite{li2022mvitv2} was introduced for image and video classification tasks. We repurpose it as a strong baseline for audio classification.

Our results demonstrate that the proposed 16-blocks with 2 views MMViT model outperforms both the AST and MMViTv2 baseline models, achieving a 32.2\% mAP on the balanced Audioset dataset and 43\% mAP on the full Audioset dataset. In comparison, our implementation of the AST reached 31.8\% on the balanced Audioset dataset and 37.2\% on the full Audioset dataset, while the MViTv2 model reached 32\% on the balanced Audioset dataset and 42.4\% on the full Audioset dataset. 

In the paper, we use audio Cutmix mentioned in section \ref{sec:dataaug} for all experiments, including baseline implementations. In addition, we perform an ablation study to evaluate the impact of audio Cutmix. In this ablation study, we only apply Mixup augmentation to half of the audio clips while leaving the other half unaltered. The findings indicate that without audio Cutmix, our model tends to overfit, resulting in a lower mAP of 39\%. Therefore, our audio Cutmix technique plays a vital role in regularizing the model and preventing overfitting.

\subsection{Image Classification Results}

\begin{table}[]
\centering
\caption{Image Classification results.}
\label{tab:image_res}
\begin{tabular}{l|l|c}
\hline
Model           & Dataset & \multicolumn{1}{l}{Top1 accuracy} \\ \hline
MViTv2          & ImageNet1K                          & 82.7                            \\
MMViT           & ImageNet1K                         & \textbf{83.2}                   \\
 \hline
\end{tabular}
\vspace{-5mm}
\end{table}

The results of our image classification experiments demonstrate comparable trends to our audio experiments. Our proposed MMViT model achieved an accuracy of 83.2\%, while the MViTv2 baseline model achieved an accuracy of 82.7\%. 

We also conducted an additional ablation study where we use 3 different views from the input. As a result, the Top-1 accuracy dropped to 82.3\%. 
We hypothesize that redundant information in all three views can negatively impact performance as model size increases. Stronger regularization methods may be needed to train model with more views. We leave this as future work.


\section{Conclusions}
In this paper, we introduced a novel multiscale Multiview Vision Transformer (MMViT) as a backbone model suitable for multiple modalities. The MMViT model combines the strengths of Multiscale Vision Transformers (MViT) and Multiview Transformers (MTV) by inputting multiple views into a multiscale stage hierarchy model. At each scale stage, a cross-attention layer is used to merge the information from views at different resolutions, thereby enabling the network to capture complex high-dimensional features. Our results demonstrate that the MMViT model outperforms state-of-the-art models in both audio and image classification tasks on public datasets.

\bibliographystyle{IEEEtran}
\bibliography{main}

\begin{thebibliography}{10}
\providecommand{\url}[1]{#1}
\csname url@samestyle\endcsname
\providecommand{\newblock}{\relax}
\providecommand{\bibinfo}[2]{#2}
\providecommand{\BIBentrySTDinterwordspacing}{\spaceskip=0pt\relax}
\providecommand{\BIBentryALTinterwordstretchfactor}{4}
\providecommand{\BIBentryALTinterwordspacing}{\spaceskip=\fontdimen2\font plus
\BIBentryALTinterwordstretchfactor\fontdimen3\font minus
  \fontdimen4\font\relax}
\providecommand{\BIBforeignlanguage}[2]{{%
\expandafter\ifx\csname l@#1\endcsname\relax
\typeout{** WARNING: IEEEtran.bst: No hyphenation pattern has been}%
\typeout{** loaded for the language `#1'. Using the pattern for}%
\typeout{** the default language instead.}%
\else
\language=\csname l@#1\endcsname
\fi
#2}}
\providecommand{\BIBdecl}{\relax}
\BIBdecl

\bibitem{he2016deep}
K.~He, X.~Zhang, S.~Ren, and J.~Sun, ``Deep residual learning for image
  recognition,'' in \emph{Proceedings of the IEEE conference on computer vision
  and pattern recognition}, 2016, pp. 770--778.

\bibitem{simonyan2014very}
K.~Simonyan and A.~Zisserman, ``Very deep convolutional networks for
  large-scale image recognition,'' \emph{arXiv preprint arXiv:1409.1556}, 2014.

\bibitem{cramer2019look}
J.~Cramer, H.-H. Wu, J.~Salamon, and J.~P. Bello, ``Look, listen, and learn
  more: Design choices for deep audio embeddings,'' in \emph{ICASSP 2019-2019
  IEEE International Conference on Acoustics, Speech and Signal Processing
  (ICASSP)}.\hskip 1em plus 0.5em minus 0.4em\relax IEEE, 2019, pp. 3852--3856.

\bibitem{arandjelovic2017look}
R.~Arandjelovic and A.~Zisserman, ``Look, listen and learn,'' in
  \emph{Proceedings of the IEEE international conference on computer vision},
  2017, pp. 609--617.

\bibitem{gong2021psla}
Y.~Gong, Y.-A. Chung, and J.~Glass, ``Psla: Improving audio tagging with
  pretraining, sampling, labeling, and aggregation,'' \emph{IEEE/ACM
  Transactions on Audio, Speech, and Language Processing}, vol.~29, pp.
  3292--3306, 2021.

\bibitem{devlin2018bert}
J.~Devlin, M.-W. Chang, K.~Lee, and K.~Toutanova, ``Bert: Pre-training of deep
  bidirectional transformers for language understanding,'' \emph{arXiv preprint
  arXiv:1810.04805}, 2018.

\bibitem{dosovitskiy2020image}
A.~Dosovitskiy, L.~Beyer, A.~Kolesnikov, D.~Weissenborn, X.~Zhai,
  T.~Unterthiner, M.~Dehghani, M.~Minderer, G.~Heigold, S.~Gelly \emph{et~al.},
  ``An image is worth 16x16 words: Transformers for image recognition at
  scale,'' \emph{arXiv preprint arXiv:2010.11929}, 2020.

\bibitem{xu2022masked}
H.~Xu, J.~Li, A.~Baevski, M.~Auli, W.~Galuba, F.~Metze, C.~Feichtenhofer
  \emph{et~al.}, ``Masked autoencoders that listen,'' \emph{arXiv preprint
  arXiv:2207.06405}, 2022.

\bibitem{chen2022beats}
S.~Chen, Y.~Wu, C.~Wang, S.~Liu, D.~Tompkins, Z.~Chen, and F.~Wei, ``Beats:
  Audio pre-training with acoustic tokenizers,'' \emph{arXiv preprint
  arXiv:2212.09058}, 2022.

\bibitem{hsu2021hubert}
W.-N. Hsu, B.~Bolte, Y.-H.~H. Tsai, K.~Lakhotia, R.~Salakhutdinov, and
  A.~Mohamed, ``Hubert: Self-supervised speech representation learning by
  masked prediction of hidden units,'' \emph{IEEE/ACM Transactions on Audio,
  Speech, and Language Processing}, vol.~29, pp. 3451--3460, 2021.

\bibitem{vaswani2017attention}
A.~Vaswani, N.~Shazeer, N.~Parmar, J.~Uszkoreit, L.~Jones, A.~N. Gomez,
  {\L}.~Kaiser, and I.~Polosukhin, ``Attention is all you need,''
  \emph{Advances in neural information processing systems}, vol.~30, 2017.

\bibitem{gong2021ast}
Y.~Gong, Y.-A. Chung, and J.~Glass, ``Ast: Audio spectrogram transformer,''
  \emph{arXiv preprint arXiv:2104.01778}, 2021.

\bibitem{fan2021multiscale}
H.~Fan, B.~Xiong, K.~Mangalam, Y.~Li, Z.~Yan, J.~Malik, and C.~Feichtenhofer,
  ``Multiscale vision transformers,'' in \emph{Proceedings of the IEEE/CVF
  International Conference on Computer Vision}, 2021, pp. 6824--6835.

\bibitem{yan2022multiview}
S.~Yan, X.~Xiong, A.~Arnab, Z.~Lu, M.~Zhang, C.~Sun, and C.~Schmid, ``Multiview
  transformers for video recognition,'' in \emph{Proceedings of the IEEE/CVF
  Conference on Computer Vision and Pattern Recognition}, 2022, pp. 3333--3343.

\bibitem{russakovsky2015imagenet}
O.~Russakovsky, J.~Deng, H.~Su, J.~Krause, S.~Satheesh, S.~Ma, Z.~Huang,
  A.~Karpathy, A.~Khosla, M.~Bernstein \emph{et~al.}, ``Imagenet large scale
  visual recognition challenge,'' \emph{International journal of computer
  vision}, vol. 115, pp. 211--252, 2015.

\bibitem{gemmeke2017audio}
J.~F. Gemmeke, D.~P. Ellis, D.~Freedman, A.~Jansen, W.~Lawrence, R.~C. Moore,
  M.~Plakal, and M.~Ritter, ``Audio set: An ontology and human-labeled dataset
  for audio events,'' in \emph{2017 IEEE international conference on acoustics,
  speech and signal processing (ICASSP)}.\hskip 1em plus 0.5em minus
  0.4em\relax IEEE, 2017, pp. 776--780.

\bibitem{he2020deberta}
P.~He, X.~Liu, J.~Gao, and W.~Chen, ``Deberta: Decoding-enhanced bert with
  disentangled attention,'' \emph{arXiv preprint arXiv:2006.03654}, 2020.

\bibitem{yang2019xlnet}
Z.~Yang, Z.~Dai, Y.~Yang, J.~Carbonell, R.~R. Salakhutdinov, and Q.~V. Le,
  ``Xlnet: Generalized autoregressive pretraining for language understanding,''
  \emph{Advances in neural information processing systems}, vol.~32, 2019.

\bibitem{radford2018improving}
A.~Radford, K.~Narasimhan, T.~Salimans, and I.~Sutskever, ``Improving language
  understanding with unsupervised learning,'' 2018.

\bibitem{touvron2021training}
H.~Touvron, M.~Cord, M.~Douze, F.~Massa, A.~Sablayrolles, and H.~J{\'e}gou,
  ``Training data-efficient image transformers \& distillation through
  attention,'' in \emph{International conference on machine learning}.\hskip
  1em plus 0.5em minus 0.4em\relax PMLR, 2021, pp. 10\,347--10\,357.

\bibitem{liu2021swin}
Z.~Liu, Y.~Lin, Y.~Cao, H.~Hu, Y.~Wei, Z.~Zhang, S.~Lin, and B.~Guo, ``Swin
  transformer: Hierarchical vision transformer using shifted windows,'' in
  \emph{Proceedings of the IEEE/CVF international conference on computer
  vision}, 2021, pp. 10\,012--10\,022.

\bibitem{koutini2021efficient}
K.~Koutini, J.~Schl{\"u}ter, H.~Eghbal-zadeh, and G.~Widmer, ``Efficient
  training of audio transformers with patchout,'' \emph{arXiv preprint
  arXiv:2110.05069}, 2021.

\bibitem{chen2022hts}
K.~Chen, X.~Du, B.~Zhu, Z.~Ma, T.~Berg-Kirkpatrick, and S.~Dubnov, ``Hts-at: A
  hierarchical token-semantic audio transformer for sound classification and
  detection,'' in \emph{ICASSP 2022-2022 IEEE International Conference on
  Acoustics, Speech and Signal Processing (ICASSP)}.\hskip 1em plus 0.5em minus
  0.4em\relax IEEE, 2022, pp. 646--650.

\bibitem{loshchilov2017decoupled}
I.~Loshchilov and F.~Hutter, ``Decoupled weight decay regularization,''
  \emph{arXiv preprint arXiv:1711.05101}, 2017.

\bibitem{zhang2018mixup}
\BIBentryALTinterwordspacing
H.~Zhang, M.~Cisse, Y.~N. Dauphin, and D.~Lopez-Paz, ``mixup: Beyond empirical
  risk minimization,'' \emph{International Conference on Learning
  Representations}, 2018. [Online]. Available:
  \url{https://openreview.net/forum?id=r1Ddp1-Rb}
\BIBentrySTDinterwordspacing

\bibitem{yun2019cutmix}
S.~Yun, D.~Han, S.~J. Oh, S.~Chun, J.~Choe, and Y.~Yoo, ``Cutmix:
  Regularization strategy to train strong classifiers with localizable
  features,'' in \emph{Proceedings of the IEEE/CVF international conference on
  computer vision}, 2019, pp. 6023--6032.

\bibitem{koutini2021receptive}
K.~Koutini, H.~Eghbal-zadeh, and G.~Widmer, ``Receptive field regularization
  techniques for audio classification and tagging with deep convolutional
  neural networks,'' \emph{IEEE/ACM Transactions on Audio, Speech, and Language
  Processing}, vol.~29, pp. 1987--2000, 2021.

\bibitem{park19e_interspeech}
D.~S. Park, W.~Chan, Y.~Zhang, C.-C. Chiu, B.~Zoph, E.~D. Cubuk, and Q.~V. Le,
  ``{SpecAugment: A Simple Data Augmentation Method for Automatic Speech
  Recognition},'' in \emph{Proc. Interspeech 2019}, 2019, pp. 2613--2617.

\bibitem{zhong2020random}
Z.~Zhong, L.~Zheng, G.~Kang, S.~Li, and Y.~Yang, ``Random erasing data
  augmentation,'' in \emph{Proceedings of the AAAI conference on artificial
  intelligence}, vol.~34, no.~07, 2020, pp. 13\,001--13\,008.

\bibitem{cubuk2020randaugment}
E.~D. Cubuk, B.~Zoph, J.~Shlens, and Q.~V. Le, ``Randaugment: Practical
  automated data augmentation with a reduced search space,'' in
  \emph{Proceedings of the IEEE/CVF conference on computer vision and pattern
  recognition workshops}, 2020, pp. 702--703.

\bibitem{li2022mvitv2}
Y.~Li, C.-Y. Wu, H.~Fan, K.~Mangalam, B.~Xiong, J.~Malik, and C.~Feichtenhofer,
  ``Mvitv2: Improved multiscale vision transformers for classification and
  detection,'' in \emph{Proceedings of the IEEE/CVF Conference on Computer
  Vision and Pattern Recognition}, 2022, pp. 4804--4814.

\end{thebibliography}

\end{document}